\documentclass[letterpaper]{article}
\usepackage{iclr2027_conference}
\iclrfinalcopy
\usepackage{times}
\usepackage[T1]{fontenc}
\usepackage[utf8]{inputenc}
\usepackage{amsmath,amssymb,bm}
\usepackage{graphicx,booktabs,multirow,makecell,tabularx,array,wrapfig}
\usepackage[font=small,labelfont=bf,skip=7pt]{caption}
\usepackage[table]{xcolor}
\usepackage{microtype,xspace,placeins,needspace}
\usepackage{algorithm,algpseudocode}
\usepackage{natbib}
\setcitestyle{authoryear,round,citesep={;},aysep={,},yysep={;}}

\usepackage{fancyhdr}
\usepackage[bottom]{footmisc}
\usepackage{hyperref,url}
\usepackage[capitalize,noabbrev]{cleveref}
\hypersetup{colorlinks=true,citecolor=blue,linkcolor=blue,urlcolor=blue,
  pdftitle={FlowHMR: Physically Plausible Motion Capture from Video}}
\newcommand{\modelname}{FlowHMR\xspace}

\newcommand{\eg}{e.g.\xspace}

\title{FlowHMR: Physically Plausible Motion Capture from Video}
\author{%
  Zhanke Wang\textsuperscript{1,*}\quad
  Chengfeng Zhao\textsuperscript{2,*}\quad
  Qing Shuai\textsuperscript{3,\textdagger}\quad
  Jingzhong Lin\textsuperscript{4}\quad
  Heng Li\textsuperscript{5}\\[3pt]
  Zeyu Ling\textsuperscript{6}\quad
  Yuxin Wen\textsuperscript{3}\quad
  Jing Li\textsuperscript{3}\quad
  Di Kang\textsuperscript{3}\quad
  Chunchao Guo\textsuperscript{3}\quad
  Linchao Bao\textsuperscript{3,\textdagger}\\[7pt]
  {\small\normalfont\textsuperscript{1}Peking University\quad
  \textsuperscript{2}The Hong Kong University of Science and Technology\quad
  \textsuperscript{3}Tencent}\\[2pt]
  {\small\normalfont\textsuperscript{4}East China Normal University\quad
  \textsuperscript{5}Sun Yat-sen University\quad
  \textsuperscript{6}Zhejiang University}%
}
\newcommand{\authornotes}{%
  \begingroup
  \renewcommand{\thefootnote}{\fnsymbol{footnote}}%
  \footnotetext[1]{Equal contribution.}%
  \footnotetext[2]{Corresponding authors.}%
  \endgroup
}
\hypersetup{pdfauthor={Zhanke Wang, Chengfeng Zhao, Qing Shuai, Jingzhong Lin, Heng Li, Zeyu Ling, Yuxin Wen, Jing Li, Di Kang, Chunchao Guo, Linchao Bao}}

\date{\small Project page: \href{https://flowhmr.github.io/}{\textcolor{red}{\nolinkurl{https://flowhmr.github.io/}}}}
\makeatletter
\renewcommand{\@maketitle}{%
  \vbox{\hsize\textwidth
    \noindent\resizebox{\textwidth}{!}{\LARGE\scshape\@title}\par
    \vskip 12pt
    \noindent\makebox[\textwidth][c]{%
      \resizebox{\textwidth}{!}{\bfseries\begin{tabular}{@{}c@{}}\@author\end{tabular}}}\par
    \vskip 7pt
    \noindent\makebox[\textwidth][c]{\@date}\par
    \vskip 12pt}}
\makeatother

\begin{document}
\maketitle
\authornotes
\begin{figure}[!ht]
  \centering
  \begin{minipage}{\linewidth}
  \centering
  \includegraphics[width=0.8\linewidth]{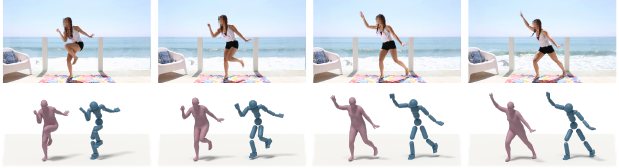}\par\vspace{6pt}
  \includegraphics[width=0.8\linewidth]{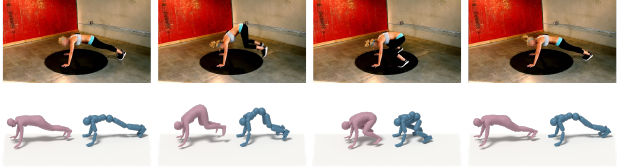}
  \end{minipage}
  \caption{\textbf{Physically plausible motion recovery.} Given a monocular video (rows 1 and 3), \modelname{} recovers global 3D human motion (pink) that a physics-based controller can successfully track in simulation (blue).}
  \label{fig:motivation}
\end{figure}

\begin{abstract}
    We present \modelname, a framework for recovering physically plausible global 3D human motion from monocular video. Previous learning-based methods typically regress human motion directly from video and train the network with geometric supervision. However, recovering human motion from monocular video is inherently ambiguous in depth, and direct regression tends to collapse toward an averaged solution. Moreover, the recovered motions are not guaranteed to be physically plausible, so physics-based tracking of them often fails. To address these challenges, we formulate video motion capture as a video-conditioned motion generation problem and first pretrain a flow matching model for this task. Given an input video, the pretrained model generates diverse motion candidates, but not all of them are faithful to the video or physically trackable. We therefore post-train the model using Group Relative Policy Optimization (GRPO) with two rewards. A fidelity reward encourages consistency with the input video. A tracking reward favors motions that a physics-based controller can track successfully. Together, these rewards shift the model's output preference, so the post-trained model stays faithful to the input video while producing more physically plausible motion. We further introduce Wild-4K, a large and diverse dataset of about 4K internet videos, for evaluating human motion recovery in the wild. Qualitative and quantitative experiments on Wild-4K show that our method outperforms state-of-the-art methods in overall motion fidelity and achieves a physical tracking success rate of 82.47\%, compared with 62.82\% for the strongest baseline, GVHMR. Code and checkpoints are available at \href{https://github.com/flowhmr/flowhmr}{\textcolor{red}{\nolinkurl{https://github.com/flowhmr/flowhmr}}}.
\end{abstract}
\clearpage
\section{Introduction}
\label{sec:introduction}

Recovering global 3D human motion from monocular video is a fundamental problem in computer vision and computer graphics~\citep{ye2023slahmr,shin2024wham,shen2024gvhmr}.
It enables a wide range of applications, from character animation to humanoid robot learning from human demonstrations~\citep{he2025asap,allshire2025videomimic}.
Internet videos capture an enormous variety of human activities, making them a rich source of motion data for these applications.
Leveraging such data requires recovering motion that is both faithful to the video and physically plausible in unconstrained, in-the-wild settings.

However, monocular motion capture is inherently challenging: depth is ambiguous, body parts are frequently occluded, and the observed image-space motion entangles human and camera movement.
Video-based methods exploit temporal context to produce coherent pose sequences~\citep{kocabas2020vibe}.
To recover motion in world coordinates, subsequent methods jointly estimate human and camera motion~\citep{ye2023slahmr}, or combine temporal modeling with camera motion and contact cues~\citep{shin2024wham,shen2024gvhmr}. 
Such learning-based methods are typically trained with geometric objectives of 2D projections, 3D poses, and global trajectories, which aligns the predicted motion with the visual observations.
Whereas, as deterministic regressors, they produce a single estimate and tend to yield over-smoothed, averaged predictions under ambiguous observations.
Generative models naturally capture this ambiguity, and generative motion priors have been incorporated into video-based motion capture~\citep{rempe2021humor,li2025genmo,wang2026duomo}.

Another limitation of geometric objectives is that they do not explicitly model the dynamics and contact of human motion.
Auxiliary losses that penalize foot skating or ground penetration impose only kinematic constraints and provide indirect supervision of the underlying dynamics.
Physics simulation complements geometric supervision by providing feedback on whether a motion is physically executable, and prior work has used physics-refined motions to further train motion generators~\citep{gillman2024scsc,li2025morph}.
For video-based motion capture, the key challenge is to incorporate such physical feedback without sacrificing fidelity to the body articulation and global trajectory observed in the input video.

We present \modelname{}, a generative framework for recovering physically plausible global 3D human motion from monocular video (Figure~\ref{fig:framework}).
We formulate video motion capture as video-conditioned motion generation and learn the conditional distribution of global human motion with flow matching~\citep{lipman2023flowmatchinggenerativemodeling}.
Our motion representation jointly encodes body pose, hand articulation, and the global root trajectory of the entire sequence in a unified world coordinate frame.
The generator is a multimodal diffusion transformer (MM-DiT) conditioned on visual features and camera information.
We pretrain the model on approximately 3,000 hours of synthetic videos paired with ground-truth 3D motion, and then post-train it via reinforcement learning (RL) with simulation-based rewards.

During post-training, we use two complementary rewards.
The fidelity reward measures agreement with the ground-truth motion, while the tracking reward measures how accurately a pretrained, frozen humanoid controller~\citep{luo2023phc,luo2024pulse} tracks the generated motion in simulation.
We adopt MixGRPO~\citep{mixgrpo2025} for policy optimization, allowing physical feedback to shape the generator's outputs while the ground-truth motion anchors reconstruction.

To evaluate motion capture in the wild, we introduce Wild-4K, a large and diverse evaluation set of approximately 4K internet videos covering everyday activities, highly dynamic movements, and severe occlusions.
We assess overall motion fidelity through blind pairwise user studies, complemented by kinematic metrics and physical tracking evaluation.
Experiments show that human evaluators prefer \modelname{} over state-of-the-art methods in terms of motion fidelity, and that \modelname{} achieves a physical tracking success rate of 82.47\%, compared with 62.82\% for the strongest baseline, GVHMR.

In summary, our main contributions are as follows:
\begin{itemize}
  \item We propose \modelname{}, a video-conditioned flow matching framework that jointly models body pose, hand articulation, and global trajectory in a world coordinate frame for motion capture from monocular video.
  \item We introduce an RL post-training scheme with fidelity and physical tracking rewards, which makes the recovered motion more physically plausible while keeping it faithful to the input video.
  \item We present Wild-4K, a large and diverse evaluation set of approximately 4K internet videos for assessing motion fidelity and physical tracking success rate.
\end{itemize}
\section{Related Work}
\label{sec:related}

\paragraph{Video-based human motion capture.}
Monocular motion capture requires inferring temporally coherent body poses and global trajectories under depth ambiguity, occlusion, and camera motion.
Optimization-based methods regularize reconstruction with learned motion priors: HuMoR~\citep{rempe2021humor} introduces a generative motion prior for robust pose estimation, while GLAMR~\citep{yuan2022glamr} and SLAHMR~\citep{ye2023slahmr} combine motion priors with optimization to recover global motion from moving-camera videos.
Feed-forward methods such as TRACE~\citep{sun2023trace}, WHAM~\citep{shin2024wham}, and GVHMR~\citep{shen2024gvhmr} directly regress world-space motion from video through temporal and geometric modeling, and PromptHMR~\citep{wang2025prompthmr} extends promptable human mesh recovery to video.
Generative models have also been applied to video-based motion capture: GENMO~\citep{li2025genmo} unifies motion estimation and generation within a single generative model, and DuoMo~\citep{wang2026duomo} performs video-conditioned camera-space reconstruction followed by world-space diffusion refinement.
Large-scale synthetic datasets such as BEDLAM~\citep{black2023bedlam} and BEDLAM2.0~\citep{teschbedlam2} provide videos paired with ground-truth body and camera annotations across diverse appearances, motions, and viewpoints.
Following this line of work, \modelname{} learns a video-conditioned distribution of world-space motion from paired video and motion data, and this generative formulation further enables post-training with physical feedback.

\paragraph{Physics-aware motion capture and generation.}
Physical constraints have been incorporated into motion capture through physics-based pose optimization in PhysCap~\citep{shimada2020physcap}, differentiable contact and stability terms in IPMAN~\citep{Tripathi2023}, and explicit dynamics modeling in D\&D~\citep{li2022dnd} and PhysPT~\citep{zhang2024physpt}.
SimPoE~\citep{yuan2021simpoe} jointly learns kinematic pose refinement and simulated character control, while PhysHMR~\citep{feng2025physhmr} learns a vision-conditioned humanoid control policy via expert distillation and reinforcement learning.
In text-to-motion generation, physics has been integrated at different stages: PhysDiff~\citep{yuan2023physdiff} projects intermediate denoised motions onto physically plausible ones via simulation; InsActor~\citep{ren2023insactor} maps diffusion-generated plans to latent skills for execution in simulation; CLoSD~\citep{tevet2025closd} feeds simulated motions back to a diffusion planner in a closed loop; and SimDiff~\citep{watanabe2025simdiff} trains environment-conditioned adapters on simulated trajectories.
In contrast, \modelname{} uses simulation only during training.
A pretrained, frozen PHC+ controller~\citep{luo2023phc,luo2024pulse} tracks the sampled motions in simulation, and its tracking performance, together with framewise agreement with the ground-truth motion, provides the reward for updating the video-conditioned generator.
At inference, the generator directly produces motion without simulation-based selection or correction.

\paragraph{Reward-based post-training of motion generators.}
Reward-based optimization has also been applied to motion generators.
ReinDiffuse~\citep{han2025reindiffuse} fine-tunes a motion diffusion model with proximal policy optimization (PPO) using rewards that penalize kinematic artifacts.
More recently, group relative policy optimization (GRPO)~\citep{shao2024deepseekmath} and its extensions to flow-based models~\citep{liu2025flowgrpo,xue2025dancegrpo} have been adopted for motion generation.
HY-Motion~\citep{hymotion2025} combines preference optimization with Flow-GRPO using semantic and kinematic rewards, and RLPF~\citep{yue2025rlpf} optimizes text-to-motion generation with GRPO using simulation-based tracking and alignment rewards.
GenTrack~\citep{ling2026gentrack} jointly optimizes motion generation and tracking through online co-training, alternating generator alignment using execution feedback with tracker updates on newly generated motions.
PhysMoDPO~\citep{zhang2026physmodpo} instead constructs preferences over generated motions from execution feedback to optimize diffusion models.
For motion capture, MotionGRPO~\citep{yao2026motiongrpo} applies GRPO to full-body motion recovery from head-mounted device signals using learned perceptual and joint-level rewards.
We build on MixGRPO~\citep{mixgrpo2025}, an efficient flow-based GRPO variant that mixes ODE and SDE sampling.
Unlike prior work, we target video-conditioned motion capture and combine a framewise fidelity reward against the ground-truth motion with a physical tracking reward.
We further analyze how these rewards affect reconstruction fidelity, physical trackability, and sample diversity.
\section{Data}
\label{sec:data}
\label{app:data}

\subsection{Synthetic Training Data}
\label{sec:paired_data}
\label{app:pretraining_data}

BEDLAM~\citep{black2023bedlam} and BEDLAM2.0~\citep{teschbedlam2} demonstrate the value of synthetic data for human motion recovery, but the available video duration remains limited: BEDLAM2.0 provides 74.52 hours.
To scale paired supervision for body, hand, and global motion recovery, we render approximately 3,000 hours of video from 650 hours of source motion, varying cameras and appearance while preserving motion annotations.

\paragraph{Motion coverage and annotation.}
\label{sec:motion_corpus}
To cover varied articulation and trajectories, we combine public motion datasets and animation assets spanning locomotion, dance, exercise, and hand motion.
We fit all SMPL-family motion data to a common SMPL-H representation~\citep{Romero2017}; other animation assets are retargeted to SMPL-H.
We remove sequences with abnormal joint positions or velocities, or foot sliding, to avoid learning artifacts from the source motions, following HY-Motion~\citep{hymotion2025}.

\begin{figure}[!t]
    \centering
    \includegraphics[width=\linewidth]{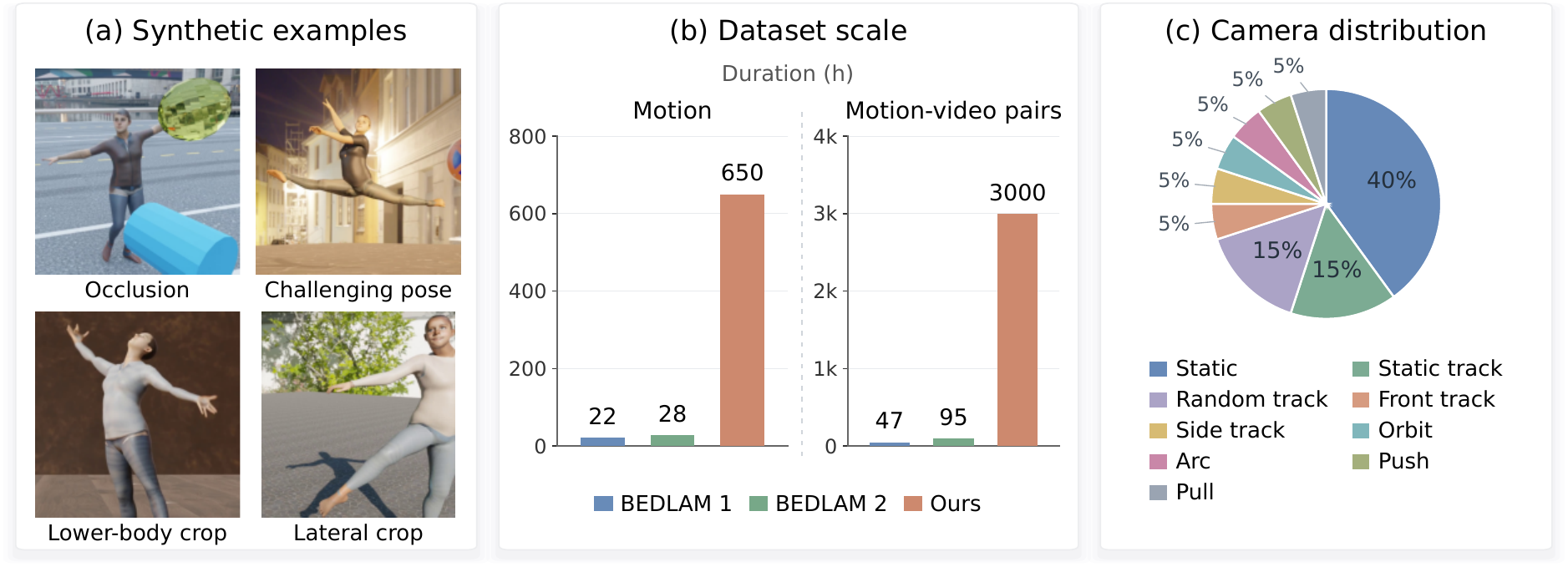}
    \caption{\textbf{Training data scale and diversity.} (a) Synthetic examples featuring occlusions, challenging poses, and partial visibility. (b) Total durations of motion data and paired motion-video data, compared with BEDLAM, the largest prior synthetic dataset. (c) Proportions of sampled camera trajectory types. We use a diverse set of camera trajectories to simulate the camera motion found in in-the-wild internet videos.}

    \label{fig:pretraining_data_examples}
\end{figure}

\paragraph{Camera variation.}
\label{app:camera_pipeline}
To distinguish human movement from apparent motion caused by the camera, we render motions with static and moving cameras across varied viewpoints, distances, and focal lengths.
Using the 3D bounding box of each motion sequence, an automated camera-generation script generates diverse camera trajectories while ensuring that the person remains in view.
Full-body views and partial image-boundary truncation expose the model to different amounts of visual evidence.

\paragraph{Appearance and occlusion.}
\label{sec:video_pairing}
\label{app:rendering_pipeline}
We source background assets from Poly Haven~\citep{polyhaven} and character textures from ATLAS~\citep{texdreamer}.
To increase visual diversity, we randomize character textures, lighting, and ground materials in Blender, and randomly place occluders in the scene foreground and in front of the camera.
Examples are shown in Figure~\ref{fig:pretraining_data_examples}(a).
Videos are rendered at 1080p and 30 fps in portrait, landscape, and square formats, with motion and camera annotations.

\paragraph{Physical tracking.}
\label{sec:physical_labels}
To identify motions that can be tracked in physics simulation, we track the annotated motions with a fixed PHC+ controller~\citep{luo2023phc,luo2024pulse}.
Requiring a tracking score above 0.6 and no early termination yields approximately 571k pairs for post-training, with the original motion annotations retained as fidelity targets.
Pretraining uses all paired data regardless of the tracking outcome.


\subsection{Wild-4K: Internet Video Evaluation}
\label{app:wild_acquisition}

Existing annotated benchmarks cover limited recording environments or numbers of sequences: Human3.6M~\citep{h36m} captures activities in an indoor laboratory, RICH~\citep{huang2022rich} records interactions in selected indoor and outdoor scenes, and 3DPW~\citep{vonMarcard2018} contains 60 video sequences.
To assess generalization across a broader range of actions, scenes, and visibility conditions, we construct Wild-4K using 4,182 internet video clips curated from Koala-36M~\citep{wang2025koala36m}.

A VLM~\citep{qwen2026qwen35} and pose/person detectors~\citep{xu2022vitpose,ge2021yolox} select clips with a single adult and full-body framing, allowing temporary occlusion.
We exclude underwater, aerial, and suspended settings to match the ground-plane simulation environment.
The VLM groups clips as \emph{normal}, \emph{difficult}, or \emph{self-occluded}, covering everyday activity, demanding actions, and occluded views.
Camera poses are estimated with VGGT-$\Omega$~\citep{wang2026vggtomega}.
Without ground-truth motion, Wild-4K supports human judgments of reconstruction quality, geometric artifact measurements, and simulation tracking, complementing the annotated benchmarks in Section~\ref{sec:experiments}.

\section{Method}
\label{sec:method}

\subsection{Overview}
\label{sec:overview}

\begin{figure}[!tp]
\centering
\includegraphics[width=\linewidth]{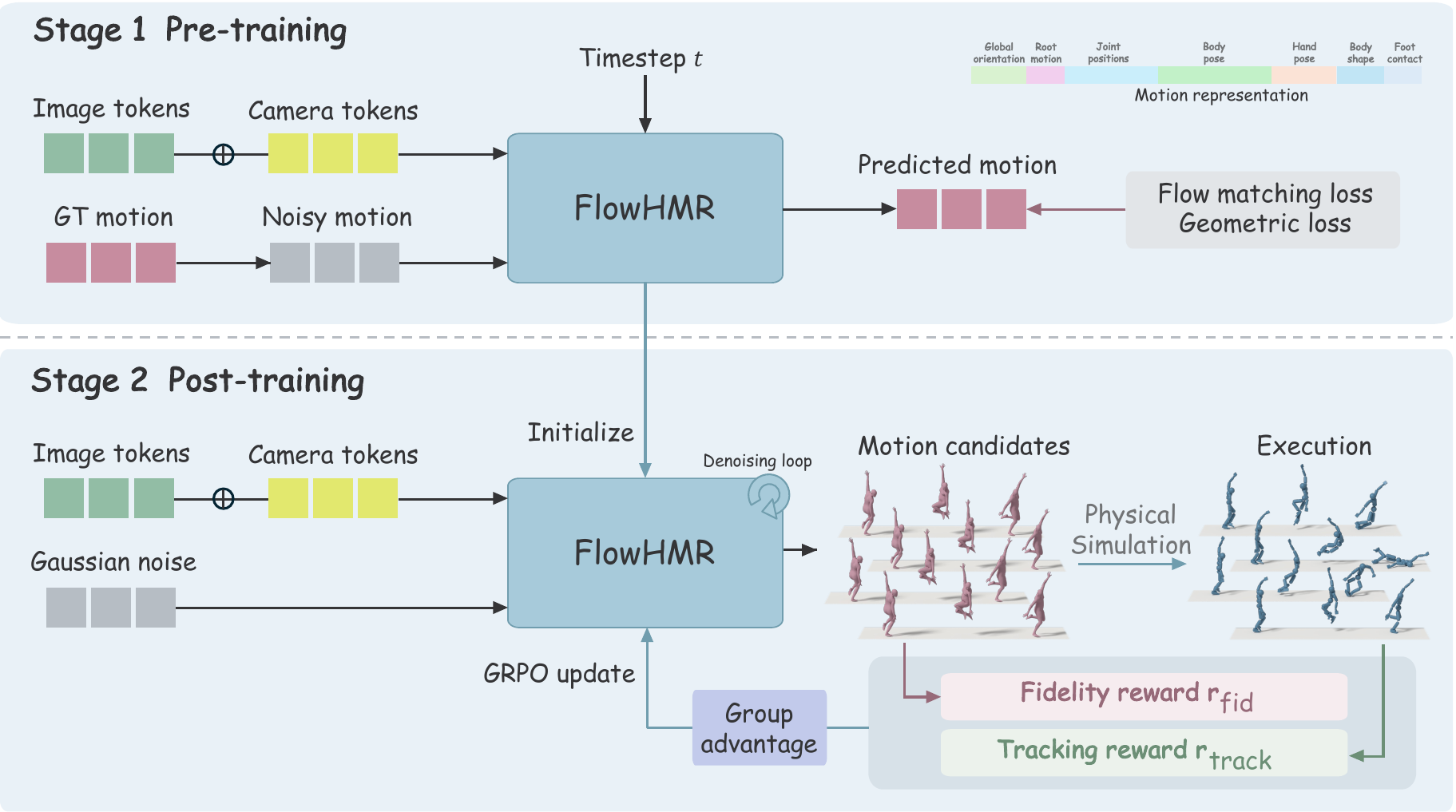}
\caption{\textbf{Overview of the \modelname{} training pipeline.} To recover faithful and physically plausible global human motion from monocular video, \modelname{} is trained in two stages. Stage 1 performs generative pretraining, in which a video-conditioned flow-matching model is trained on paired video--motion data with a flow-matching loss and a forward kinematics loss. Stage 2 performs physics-based post-training, in which the model, initialized from the Stage 1 checkpoint, is further fine-tuned with GRPO using two rewards: a fidelity reward that measures agreement with the ground-truth motion, and a physical tracking reward that measures how accurately a frozen, pretrained controller tracks the generated motion in a physics simulator.}

\label{fig:framework}
\end{figure}

Given a monocular video, \modelname{} recovers global 3D human motion that is faithful to the video and physically plausible.
Figure~\ref{fig:framework} illustrates the overall framework, which consists of two stages.
In the first stage (Section~\ref{sec:pretraining}), we pretrain a video-conditioned flow matching model on the paired video and motion data described in Section~\ref{sec:data}.
In the second stage (Section~\ref{sec:posttraining}), we post-train the model via reinforcement learning with a fidelity reward and a physical tracking reward.

\subsection{Generative Video-to-Motion Pretraining}
\label{sec:pretraining}

Depth ambiguity and occlusion make monocular motion capture ill-posed: multiple 3D motions can explain the same visual observations.
We therefore formulate it as conditional generation and use flow matching to learn the distribution of global human motion conditioned on video and camera observations.

\subsubsection{Conditional Motion Model}
\label{sec:conditional_motion_model}
\label{sec:motion_representation}
\label{sec:conditioning}

\paragraph{Motion representation.}
Recovering global motion from a moving camera requires a coordinate frame that is consistent across frames.
We express each sequence in a gravity-aligned world frame whose heading is defined by the camera orientation at the first frame, and translate the motion horizontally so that it starts at the origin.
An $N$-frame motion is encoded as $\bm{x}\in\mathbb{R}^{N\times D}$.
Following Kimodo~\citep{Kimodo2026} and related motion representations~\citep{Guo2022,hymotion2025}, we define the root as the temporally smoothed pelvis trajectory.
Each frame contains
(i) the horizontal velocity and absolute height of the root;
(ii) the SMPL-H~\citep{Romero2017} joint positions, each decomposed into a horizontal offset relative to the root and an absolute height;
(iii) the joint rotations in the continuous 6D representation~\citep{Zhou2019}, where the pelvis orientation is expressed in the world frame and all other rotations are relative to their parent joints; and
(iv) the body shape parameters and foot contact labels.
All features are standardized using statistics computed on the training set.
This representation jointly encodes body and hand articulation together with the global trajectory.

\paragraph{Conditioning.}
The model is conditioned on per-frame visual features and camera orientations in the world frame.
For each frame, we crop the person region and extract visual features using the frozen, pose-aware image encoder of SAM 3D Body~\citep{Yang2025SAM3DBody}.
The visual features and camera rotations are separately projected to a common dimension and summed to form per-frame condition tokens, denoted by $\bm{c}$.

\paragraph{Network architecture.}
We adopt the 0.46B-parameter multimodal diffusion transformer (MM-DiT) architecture of HY-Motion~\citep{hymotion2025}.
We replace its text-conditioning branch with a visual-conditioning branch whose tokens are aligned frame by frame with the motion sequence.
At flow time $t$, the noisy motion $\bm{x}_t$ is embedded into per-frame motion tokens.
The network first applies dual-stream blocks, which use separate projection weights for motion and condition tokens, and then single-stream blocks, which process the concatenated token sequence with shared weights.
The time $t$ is embedded and modulates each block through adaptive layer normalization (adaLN), and a final linear layer predicts the clean motion $\hat{\bm{x}}_1=f_\theta(\bm{x}_t,\bm{c},t)$.
In both types of blocks, we apply local temporal attention: each motion token attends to motion and condition tokens within a local temporal window, whereas each condition token attends only to condition tokens within the window.
This design aggregates evidence across neighboring frames while keeping the condition features independent of the noisy motion.

\Needspace{14\baselineskip}
\subsubsection{Pretraining with Flow Matching}
\label{sec:pretraining_objectives}
\label{sec:motion_generation}

We train the conditional generator with flow matching~\citep{lipman2023flowmatchinggenerativemodeling} on paired videos and ground-truth 3D motion, learning to transport Gaussian noise to human motion.
For each training pair, let $\bm{x}_1$ denote the ground-truth motion and $\bm{c}$ the corresponding conditions.
We sample Gaussian noise $\bm{x}_0$ and a time $t$, and linearly interpolate between $\bm{x}_0$ and $\bm{x}_1$ to obtain the noisy motion $\bm{x}_t$.
Following JiT~\citep{li2025jit}, the network directly predicts the clean motion, which we convert into a velocity prediction:
\begin{equation}
\begin{split}
    \bm{x}_t&=(1-t)\bm{x}_0+t\bm{x}_1, \qquad \bm{x}_0\sim\mathcal{N}(\bm{0},\bm{I}),\\
    \hat{\bm{v}}_\theta&=\frac{f_\theta(\bm{x}_t,\bm{c},t)-\bm{x}_t}{d(t)},
    \qquad \bm{v}^{\ast}=\frac{\bm{x}_1-\bm{x}_t}{d(t)},
\end{split}
\label{eq:pretrain_fm}
\end{equation}
where $t\in[0,1]$ is sampled from a logit-normal distribution~\citep{Esser2024}, and $d(t)=\max(1-t,\epsilon)$ clips the denominator near $t=1$ for numerical stability.
The flow matching loss is a weighted sum of Smooth-L1 losses computed separately for each feature group:
\begin{equation}
    \mathcal{L}_{\mathrm{FM}}=\sum_{k} w_k\,\ell_{\mathrm{SL1}}\big(\hat{\bm{v}}_\theta^{(k)},\bm{v}^{\ast(k)}\big),
\label{eq:fm_loss}
\end{equation}
where $k$ indexes the root trajectory, articulation, body shape, and foot contact features, and $w_k$ balances their contributions.
To directly supervise the global body configuration, we further decode both $\hat{\bm{x}}_1$ and $\bm{x}_1$ into global joint rotations and positions via SMPL-H forward kinematics (FK), each using its own body shape, and apply Smooth-L1 losses between them without pelvis or Procrustes alignment.
The total pretraining objective is
\begin{equation}
    \mathcal{L}_{\mathrm{pre}}
    =\mathcal{L}_{\mathrm{FM}}
    +\lambda_{\mathrm{rot}}\mathcal{L}_{\mathrm{rot}}
    +\lambda_{\mathrm{pos}}\mathcal{L}_{\mathrm{pos}}.
\label{eq:pretrain_total}
\end{equation}
Since no alignment is applied, $\mathcal{L}_{\mathrm{rot}}$ and $\mathcal{L}_{\mathrm{pos}}$ penalize errors in both the global trajectory and the articulation.
All losses are averaged over feature dimensions and valid frames, excluding padded frames.

\paragraph{Implementation details.}
We pretrain the model for 500K steps with a global batch size of 1,024, using AdamW with a learning rate cosine-decayed from $10^{-4}$ to $10^{-5}$.
During training, we randomly drop the visual conditions while always retaining the camera conditions.
This encourages the model to also learn motion generation without visual conditioning, rather than overfitting to the video conditions.

\paragraph{Inference.}
At inference, we integrate the learned velocity field from Gaussian noise using 20 uniform Euler steps, without classifier-free guidance (CFG).
To decode the global motion, we integrate the predicted root velocity to obtain the horizontal root trajectory, and add the predicted horizontal offset of the pelvis joint to obtain the horizontal pelvis position.
The vertical pelvis position is given by the predicted absolute height of the pelvis joint; the root height feature is supervised during training but not used for decoding.
We average the predicted shape parameters over time to obtain a single body shape per sequence, and compute the SMPL-H global translation by subtracting the shape-dependent rest-pose pelvis offset from the decoded pelvis position.
The final SMPL-H body is then determined by the predicted joint rotations, body shape, and global translation.

\subsection{Physics-Based Post-Training}
\label{sec:posttraining}
\label{sec:reward_motivation}

The pretraining objectives supervise geometric reconstruction but do not explicitly account for dynamics and contact.
We therefore post-train the generator with feedback from physics simulation.
A pretrained, frozen humanoid controller tracks each generated motion in simulation, and its tracking accuracy serves as a reward for reinforcement learning.
Since optimizing tracking alone may favor motions that are easy to execute but deviate from the input video, we combine it with a fidelity reward that measures agreement with the ground-truth motion.

\subsubsection{Rewards}
\label{sec:posttraining_rewards}

Given the conditions $\bm{c}$ of a training video and its ground-truth motion $\bm{x}^{\mathrm{gt}}$, we sample a group of $G$ motions $\{\bm{x}^{(i)}\}_{i=1}^{G}$ using the stochastic sampler described in Section~\ref{sec:grpo}.
Each sampled motion is evaluated with a fidelity reward and a physical tracking reward.
For brevity, we omit the sample index $i$ below.

\paragraph{Fidelity reward.}
The fidelity reward measures how closely a sampled motion matches the ground-truth motion.
We decode both motions into SMPL-H joint positions via FK, denoted by $\bm{p}_{n,j}$ and $\bm{p}^{\mathrm{gt}}_{n,j}$ for frame $n$ and joint $j$, and define
\begin{equation}
    r_{\mathrm{fid}}
    =\exp\!\left[-\frac{1}{\tau NJ}\sum_{n=1}^{N}\sum_{j=1}^{J}
    \left\|\bm{p}_{n,j}-\bm{p}^{\mathrm{gt}}_{n,j}\right\|_2\right],
    \label{eq:posttrain_reconstruction_reward}
\end{equation}
where $N$ is the number of valid frames, $J$ is the number of body and hand joints, and $\tau$ is a temperature that controls the reward scale.
As with the FK losses in pretraining, joint positions are compared in the world frame without any alignment, so the reward accounts for errors in both articulation and global trajectory.

\paragraph{Physical tracking reward.}
The tracking reward measures whether a sampled motion can be executed under physical dynamics and contact.
We use a pretrained, frozen PHC+ controller~\citep{luo2023phc,luo2024pulse} to track the sampled motion in Isaac Gym~\citep{makoviychuk2021isaacgym}, using the sampled motion itself as the tracking reference.
Let $\bm{b}^{\mathrm{ref}}_{n,m}$ and $\bm{b}^{\mathrm{sim}}_{n,m}$ denote the positions of rigid body $m$ at step $n$ in the reference motion and in the simulated rollout, respectively.
The tracking reward is
\begin{equation}
    r_{\mathrm{track}}
    =\exp\!\left[-\frac{1}{\tau H}\sum_{n=1}^{H}\min\!\left(
    \frac{1}{M}\sum_{m=1}^{M}
    \left\|\bm{b}^{\mathrm{sim}}_{n,m}-\bm{b}^{\mathrm{ref}}_{n,m}\right\|_2,\;
    d_{\max}\right)\right],
    \label{eq:posttrain_tracking_reward}
\end{equation}
where $H=N-1$ is the number of simulated steps after initialization, $M$ is the number of rigid bodies of the simulated humanoid, and $d_{\max}$ caps the per-step error, \eg, after the humanoid falls.
We set $\tau=0.15\,\mathrm{m}$ for both rewards and $d_{\max}=0.5\,\mathrm{m}$.
Following PHC~\citep{luo2023phc}, a rollout is considered a failure if the mean body position error exceeds $0.5\,\mathrm{m}$ at any step, and a success otherwise.

Before simulation, each motion is converted to the coordinate convention of the controller at 30\,Hz with upright-start alignment, and the humanoid is initialized at the first reference frame.
We disable the fail-state recovery and motion looping of the controller, so that tracking failures are reflected in the reward rather than corrected by the controller.
Motions whose conversion or simulation fails receive a tracking reward of zero.

\subsubsection{Reward-Guided Policy Optimization}
\label{sec:grpo}

Since the simulator is non-differentiable and both rewards are defined only on complete motions, we optimize the generator with policy gradients instead of backpropagating through the simulation.
We initialize the generator from the pretrained model and adopt MixGRPO~\citep{mixgrpo2025}, referred to as GRPO in our experiments.

\paragraph{Group sampling.}
For each training video, the current generator samples a group of $G=8$ motions.
Following MixGRPO, stochastic (SDE) sampling is applied only within a window of denoising steps, and deterministic Euler (ODE) steps are used elsewhere.
All motions in a group share the same initial Gaussian noise and differ only through the independent noise injected within the stochastic window, so that differences in their rewards can be attributed to the transitions being optimized.

\paragraph{Advantage estimation.}
We keep only groups that contain both successful and failed tracking outcomes, since groups in which all motions succeed or all fail provide little contrast between trackable and untrackable motions.
Within each retained group, the two rewards are standardized separately, so that neither reward dominates due to differences in scale, and then combined with equal weights:
\begin{equation}
    A^{(i)}=\operatorname{clip}\!\left(
    \frac{r_{\mathrm{fid}}^{(i)}-\mu_{\mathrm{fid}}}{\sigma_{\mathrm{fid}}}
    +\frac{r_{\mathrm{track}}^{(i)}-\mu_{\mathrm{track}}}{\sigma_{\mathrm{track}}},
    \;-A_{\max},\;A_{\max}\right),
    \label{eq:advantage}
\end{equation}
where $\mu$ and $\sigma$ denote the mean and standard deviation of the corresponding reward within the group.
The advantages weight the policy updates only at the stochastic transitions within the sampling window.
Only the generator is updated; the visual encoder and the controller remain frozen.
\FloatBarrier
\section{Experiments}
\label{sec:experiments}

We evaluate motion capture from internet videos through blinded human comparisons, supported by simulation tracking and geometric measurements.

\subsection{Experimental Setup}
\label{sec:setup}

\paragraph{Datasets.}
Training data and Wild-4K construction are described in Section~\ref{sec:data}. Wild-4K contains 4,182 internet video clips without ground-truth motion annotations.
We use the full set for tracking and geometric evaluation.
For human evaluation, we sample comparison pairs without replacement from a pool constructed using 500 Wild-4K videos; the evaluated video sets differ across baselines.
RICH~\citep{huang2022rich} provides annotated reconstruction and physical-execution comparisons on 185 sequences (Section~\ref{app:rich_results}). The internal Studio set supports pretraining ablations, as described in Section~\ref{sec:pretraining_ablation}.

\paragraph{Human evaluation.}
Our primary evaluation uses blinded pairwise comparisons of direct predictions for the same video, with randomized A/B order; each sampled pair is judged once by a single annotator.
Annotators judge overall motion quality through naturalness, video fidelity, global trajectory, and ground-contact artifacts.
Finger poses use a shared default and are excluded from assessment.
G/S/B denote wins, ties, and losses for our method, aggregated over valid judgments.

\paragraph{Evaluation metrics.}
To complement the overall human judgment, we report \emph{Success Rate} under a fixed PHC+ controller, with \emph{Tracking Score} additionally reported in the post-training ablations. Success requires $r_{\mathrm{track}}>0.6$, no early termination, and completion of the scored horizon; all evaluated clips enter the denominator.
Following PhysDiff~\citep{yuan2023physdiff}, we measure \emph{Penetration}, \emph{Floating}, and \emph{Skating} on generated motions before simulation.
Penetration and Floating use the lowest mesh vertex's distance below and above the ground with a 5\,mm tolerance; Skating measures horizontal foot displacement during contact. We average each error within a sequence and report its median over all 4,182 clips in millimeters.

\paragraph{Compared methods.}
We compare with GVHMR, PromptHMR, GENMO, and DuoMo~\citep{shen2024gvhmr,wang2025prompthmr,li2025genmo,wang2026duomo}.
\emph{Ours w/o RL} and \emph{Ours} denote \modelname{} before and after post-training.

\paragraph{Inference settings.}
\modelname{} generates one motion per video with seed 0 and 20 sampling steps, without classifier-free guidance, simulation-based selection, or motion correction.
Wild-4K tracking evaluation uses at most 360 frames per clip at 30\,Hz.

\subsection{Results on Internet Videos}
\label{sec:quantitative_results}

\begin{table}[!htbp]
\centering\small
\caption{\textbf{Quantitative Results on Wild-4K.} \textbf{Left:} our method outperforms all baselines in blinded human evaluation, with more wins than losses in every comparison. Bars show win/tie/loss percentages over valid judgments; evaluated videos differ across comparisons. \textbf{Right:} simulation Success Rate (SR, \%) under PHC+ and physical errors on 4,182 clips. Our method achieves the highest SR, the lowest penetration (Pen.) and floating (Float) errors, and the second-lowest skating (Skate) error. }
\label{tab:main_results}
\newsavebox{\internetmetricstable}
\sbox{\internetmetricstable}{%
  \begin{minipage}{0.49\linewidth}
  \centering\small\setlength{\tabcolsep}{2pt}
\renewcommand{\arraystretch}{1.15}
\begin{tabular*}{\linewidth}{@{\extracolsep{\fill}}l*{4}{r}@{}}
\toprule
Method & SR\,$\uparrow$ & Pen.\,$\downarrow$ & Float\,$\downarrow$ & Skate\,$\downarrow$ \\
\midrule
GVHMR & 62.82 & 1.02 & 0.81 & \textbf{1.59} \\
PromptHMR & 60.33 & 1.10 & 2.71 & 3.64 \\
GENMO & 58.80 & 1.52 & 1.20 & 2.55 \\
DuoMo & 56.71 & 1.76 & 2.08 & 2.56 \\
\midrule
Ours w/o RL & \underline{78.79} & \underline{0.60} & \underline{0.53} & 2.48 \\
Ours & \textbf{82.47} & \textbf{0.06} & \textbf{0.20} & \underline{2.17} \\
\bottomrule
\end{tabular*}

  \end{minipage}}
\medskip
\begin{minipage}[c]{0.49\linewidth}
\centering
\includegraphics[width=\linewidth,height=\dimexpr\ht\internetmetricstable+\dp\internetmetricstable\relax]{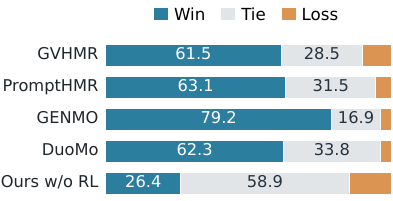}
\end{minipage}\hfill
\usebox{\internetmetricstable}
\end{table}

\subsubsection{Blinded Human Evaluation}
\label{sec:human_evaluation}

In \cref{tab:main_results} (left), \modelname{} wins 61.5--79.2\% of valid annotations against the four external methods, with losses of 3.8--10.0\%.

\Needspace{5\baselineskip}
The comparison with Ours w/o RL evaluates the perceptual effect of post-training within the same model.
Among 129 valid annotations, the post-trained model receives 34 wins (26.4\%), 76 ties (58.9\%), and 19 losses (14.7\%).
Post-training is preferred more often than the pretrained model, while the majority of comparisons are judged equal.

\subsubsection{Tracking and Physical Plausibility}
\label{sec:main_results}

On Wild-4K (\cref{tab:main_results}, right), Ours w/o RL reaches a Success Rate of 78.79\%, compared with the highest external baseline rate of 62.82\%.
Post-training raises this rate to 82.47\%, a gain of 3.68 percentage points.

To determine whether the tracking gains are accompanied by fewer geometric defects, we examine the generated motions before simulation.
Median Penetration decreases from 0.60 to 0.06\,mm, Floating from 0.53 to 0.20\,mm, and Skating from 2.48 to 2.17\,mm after post-training.
GVHMR~\citep{shen2024gvhmr} still has lower Skating at 1.59\,mm.

\begin{figure}[!tp]
\centering
\includegraphics[width=\linewidth]{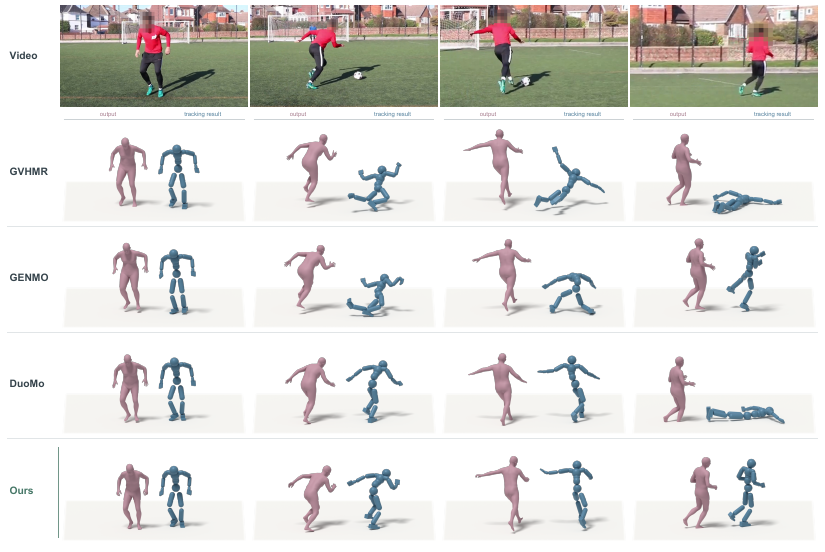}
\caption{\textbf{Qualitative comparison of reconstruction and simulation tracking.} The top row shows the input video, followed by GVHMR~\citep{shen2024gvhmr}, GENMO~\citep{li2025genmo}, DuoMo~\citep{wang2026duomo}, and \modelname{}. Each method panel pairs the reconstructed output (pink, left) with its tracking result in simulation (blue, right). Our method enables successful simulation tracking of this sequence, whereas all compared baselines fail.}
\label{fig:qualitative}
\end{figure}

\Needspace{8\baselineskip}
\subsection{Reconstruction and Physical Execution on RICH}
\label{app:rich_results}
\label{sec:reconstruction_artifacts}

Wild-4K measures recovery quality on internet videos without motion annotations. To evaluate reconstruction accuracy together with physical execution on the same annotated sequences, we additionally compare methods on RICH~\citep{huang2022rich}. Table~\ref{tab:reconstruction_results} reports geometric errors and PHC+ tracking for all 185 sequences, including tracking failures. WA-/W-/PA-MPJPE measure world-aligned, world-coordinate, and Procrustes-aligned joint position errors, respectively. RTE denotes root translation error; RTE and Jitter are evaluated over complete sequences.

\begin{table}[!htbp]
\centering\small\setlength{\tabcolsep}{4pt}
\caption{\textbf{Quantitative reconstruction and physical-execution results on RICH.} WA/W/PA denote WA-/W-/PA-MPJPE; these and foot sliding (FS) use mm. RTE is in \%; Jitter in $\mathrm{m/s^3}$. PHC+ Success (\%) and tracking MPJPE (mm) use all 185 sequences, including failures; the latter compares execution with the generated reference motion. Ours achieves the highest PHC+ Success Rate and the lowest tracking MPJPE, and competitive reconstruction results.}
\label{tab:reconstruction_results}
\begin{tabular*}{\linewidth}{@{\extracolsep{\fill}}l*{8}{r}@{}}
\toprule
\multirow{2}{*}{Method} & \multicolumn{4}{c}{Reconstruction} & \multicolumn{2}{c}{Motion quality} & \multicolumn{2}{c}{PHC+ execution} \\
\cmidrule(lr){2-5}\cmidrule(lr){6-7}\cmidrule(lr){8-9}
 & WA\,$\downarrow$ & W\,$\downarrow$ & PA\,$\downarrow$ & RTE\,$\downarrow$ & Jitter\,$\downarrow$ & FS\,$\downarrow$ & Success\,$\uparrow$ & MPJPE\,$\downarrow$ \\
\midrule
GVHMR & 82.83 & 133.08 & 40.05 & 2.49 & 20.69 & 3.64 & 28.11 & 154.3 \\
PromptHMR & 72.25 & 115.55 & 41.92 & \underline{1.71} & 19.56 & 4.27 & 35.14 & 150.5 \\
GENMO & 88.12 & 140.51 & 42.20 & 2.49 & 16.44 & 4.02 & 35.68 & 148.2 \\
DuoMo & \textbf{52.86} & \textbf{78.28} & 37.43 & \textbf{1.33} & 7.53 & \underline{3.29} & 24.86 & 179.1 \\
\midrule
Ours w/o RL & \underline{55.70} & \underline{88.37} & \textbf{30.78} & 1.92 & \underline{7.07} & 3.63 & \underline{48.65} & \underline{133.5} \\
Ours & 56.68 & 88.43 & \underline{31.40} & 1.82 & \textbf{6.71} & \textbf{3.14} & \textbf{51.35} & \textbf{118.8} \\
\bottomrule
\end{tabular*}
\end{table}

Post-training increases \modelname{}'s PHC+ Success Rate from 48.65\% to 51.35\% and reduces tracking MPJPE from 133.5 to 118.8\,mm. The final model has the highest success rate and lowest tracking error among the evaluated methods. Its WA-MPJPE and PA-MPJPE increase slightly relative to the pretrained model, from 55.70 to 56.68\,mm and from 30.78 to 31.40\,mm, respectively, while Jitter and Foot-sliding decrease. DuoMo~\citep{wang2026duomo} has lower WA-MPJPE, W-MPJPE, and RTE, but a lower PHC+ Success Rate. Reconstruction accuracy and controller tracking therefore provide complementary evidence about the recovered motions.

\FloatBarrier
\subsection{Execution under SONIC}
\label{app:sonic}

\begin{figure}[!tp]
  \centering
  \includegraphics[width=\linewidth]{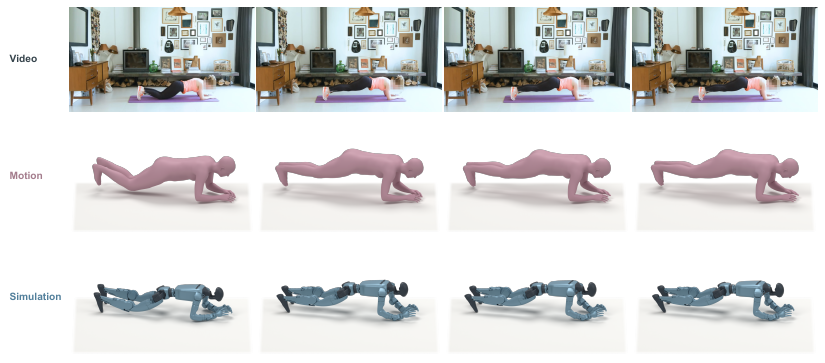}
  \caption{\textbf{SONIC execution on Unitree G1.} Rows show the input video, the motion reconstructed by \modelname{} (pink), and SONIC tracking in Isaac Lab (blue).}
  \label{fig:sonic_qualitative}
\end{figure}

To evaluate physical execution under a controller not used in post-training, we retarget Wild-4K reconstructions to the Unitree G1 humanoid with GMR~\citep{joao2025gmr} and track them with SONIC~\citep{luo2026sonic} in Isaac Lab~\citep{mittal2025isaaclab}, counting a clip as successful if the reference sequence finishes without a non-timeout termination. We evaluate on the 615 clips on which at least one method succeeds. As shown in Table~\ref{tab:sonic_results}, \modelname{} achieves the highest success rate of 77.72\%, outperforming the strongest baseline, GVHMR~\citep{shen2024gvhmr} (54.15\%). Figure~\ref{fig:sonic_qualitative} shows an example of \modelname{} motion successfully executed by SONIC.

\begin{table}[!htbp]
\centering\small
\caption{\textbf{SONIC success rate on Wild-4K.} Bold and underlining mark the best and second-best values.}
\label{tab:sonic_results}
\setlength{\tabcolsep}{2pt}
\begin{tabular}{@{}lr@{}}
\toprule
Method & Success rate (\%) $\uparrow$ \\
\midrule
PromptHMR & 44.07 \\
GENMO & 48.78 \\
DuoMo & 51.06 \\
GVHMR & \underline{54.15} \\
\modelname{} (ours) & \textbf{77.72} \\
\bottomrule
\end{tabular}
\end{table}

\FloatBarrier
\subsection{Ablation Studies}
\label{sec:ablations}

\subsubsection{Pretraining Design}
\label{sec:pretraining_ablation}

We study how training data, generative modeling, and model size affect reconstruction on the internal Studio set.
Studio contains 300 video sequences spanning diverse motion categories with optical motion capture ground truth, supporting quantitative comparisons of body, hand, and trajectory reconstruction.
\Cref{tab:pretraining_ablation} reports mean results across its motion categories; MPJRE$_{\mathrm{hand}}$ denotes mean per-joint hand rotation error.

\begin{table}[!htbp]
\centering\small
\setlength{\tabcolsep}{2pt}
\renewcommand{\arraystretch}{1.08}
\caption{\textbf{Ablations of the pretrained model.}}
\label{tab:pretraining_ablation}
\label{legacy:tab:pretraining_ablation}
\label{legacy:tab:data_ablation}
\label{legacy:tab:model_size_ablation}
\begin{tabular*}{\linewidth}{@{\extracolsep{\fill}}ll*{6}{r}@{}}
\toprule
Factor & Setting & \makecell{WA-MPJPE\\(mm) $\downarrow$} & \makecell{MPJPE\\(mm) $\downarrow$} & \makecell{RTE\\(mm) $\downarrow$} & \makecell{MPJRE$_{\mathrm{hand}}$\\($^\circ$) $\downarrow$} & \makecell{Jitter\\(mm/s$^2$) $\downarrow$} & \makecell{Foot-sliding\\(mm) $\downarrow$} \\
\midrule
Data scale & 200 h & \underline{149.44} & \underline{90.92} & \underline{4.91} & \underline{25.17} & \underline{16.34} & \underline{3.53} \\
 & 3,000 h & \textbf{67.82} & \textbf{43.49} & \textbf{1.86} & \textbf{16.08} & \textbf{11.75} & \textbf{2.55} \\
\midrule
Prediction & Regression & \underline{73.28} & \textbf{43.44} & \underline{2.39} & \textbf{14.55} & \underline{18.01} & \underline{3.94} \\
 & Flow Matching & \textbf{67.82} & \underline{43.49} & \textbf{1.86} & \underline{16.08} & \textbf{11.75} & \textbf{2.55} \\
\midrule
Model size & 0.2B & 67.82 & 43.49 & 1.86 & \underline{16.08} & 11.75 & 2.55 \\
 & 0.46B & \underline{63.39} & \textbf{42.31} & \textbf{1.81} & 16.48 & \underline{11.03} & \textbf{2.27} \\
 & 1B & \textbf{62.84} & \underline{42.37} & \underline{1.83} & \textbf{15.96} & \textbf{10.93} & \underline{2.28} \\
\bottomrule
\end{tabular*}
\end{table}

\paragraph{Training data scale.}
We evaluate the effect of pretraining data scale by comparing models trained on 200 and 3,000 hours of pretraining data.
The 3,000-hour model improves all six reported metrics.
WA-MPJPE decreases from 149.44 to 67.82\,mm and hand-rotation error from 25.17 to 16.08 degrees.

\paragraph{Generative formulation.}
We compare Flow Matching with feed-forward regression to assess the effect of generative modeling on reconstruction.
Flow Matching reduces WA-MPJPE from 73.28 to 67.82\,mm and also improves RTE, Jitter, and Foot-sliding.
MPJPE is similar (43.49 versus 43.44\,mm), while hand-rotation error is higher (16.08 versus 14.55 degrees).

\paragraph{Model size.}
Increasing model size from 0.2B to 1B reduces WA-MPJPE from 67.82 to 62.84\,mm and Jitter from 11.75 to 10.93.
The gains are not monotonic across all metrics: the 0.46B model has the lowest MPJPE, RTE, and Foot-sliding among the three sizes.

\Needspace{22\baselineskip}
\subsubsection{Post-training Design}
\label{sec:comparison}

\begin{table}[!htbp]
\centering\small
\setlength{\tabcolsep}{3pt}
\renewcommand{\arraystretch}{1.08}
\caption{\textbf{Ablation of post-training.} We compare different post-training methods and reinforcement-learning reward settings. Our GRPO with fidelity (Fid.) and physical tracking (Track.) rewards achieves higher simulation success than both SFT variants while keeping reconstruction errors close to the pretrained model. Optimizing tracking alone yields high success at the cost of severe reconstruction errors. Adding a reward based on Phys-Err (penetration, floating,
and skating errors)~\citep{yuan2023physdiff} yields comparable
tracking success but worse reconstruction accuracy than
our joint-reward design. $^{*}$ marks reward hacking, excluded from ranking.}
\label{tab:posttraining_ablation}
\label{tab:simulation_roles}
\label{tab:reward_ablation}
\begin{tabular*}{\linewidth}{@{\extracolsep{\fill}}llrrrr@{\hspace{0.5em}}}
\toprule
 & & \multicolumn{2}{c}{RICH} & \multicolumn{2}{c}{Wild-4K (PHC+)} \\
\cmidrule(lr){3-4}\cmidrule(lr){5-6}
Method & Target / reward & \makecell{WA-MPJPE\\(mm) $\downarrow$} & \makecell{PA-MPJPE\\(mm) $\downarrow$} & \makecell{Success\\(\%) $\uparrow$} & Score $\uparrow$ \\
\midrule
Pretrained & \textemdash & \underline{55.70} & \textbf{30.78} & 78.79 & 0.692 \\
\midrule
SFT & Filtered originals & \textbf{54.80} & \underline{30.88} & 78.38 & 0.690 \\
 & Projected targets & 70.80 & 42.13 & 80.25 & 0.706 \\
\midrule
GRPO & Fid. only & 57.29 & 31.62 & 73.79 & 0.658 \\
 & Track. only$^{*}$ & 255.67 & 139.20 & 99.62\rlap{$^{*}$} & 0.942\rlap{$^{*}$} \\
 & Fid. + Phys-Err & 62.15 & 37.43 & 77.98 & 0.671 \\
 & Fid. + Track. + Phys-Err & 60.66 & 34.38 & \textbf{82.57} & \underline{0.744} \\
\textbf{Ours} & Fid. + Track. & 56.68 & 31.40 & \underline{82.47} & \textbf{0.746} \\
\bottomrule
\end{tabular*}
\end{table}

\Cref{tab:posttraining_ablation} compares supervision targets and reward combinations using the pretrained model as a shared reference.
Reconstruction and tracking are reported together to assess whether tracking gains preserve the recovered action.

\paragraph{Post-training strategies.}
We test whether supervised fine-tuning (SFT) with filtered original targets or simulated trajectories obtains tracking gains comparable to joint-reward GRPO.
SFT with filtered originals retains the original motion targets and reduces RICH WA-MPJPE to 54.80\,mm, but does not improve Wild-4K tracking over the pretrained model.
Projected-target SFT uses simulated body poses and root trajectories as supervision targets.
It increases Success Rate to 80.25\%, but raises WA-MPJPE to 70.80\,mm and PA-MPJPE to 42.13\,mm.
Joint-reward GRPO reaches 82.47\% Success Rate with smaller reconstruction costs, retaining the original annotation as its fidelity reference.

\paragraph{Reward components.}
The fidelity and physical tracking rewards serve as different objectives, so we first test what happens when either is used alone.
Fidelity-only GRPO does not improve reconstruction or tracking over the pretrained model.
Tracking-only GRPO reaches 99.62\% Success Rate, but WA-MPJPE and PA-MPJPE increase to 255.67 and 139.20\,mm.
This reward-hacking behavior shows why high tracking scores must be assessed together with reconstruction fidelity.

We also evaluate a reward based on Phys-Err (penetration,
floating, and skating errors)~\citep{yuan2023physdiff}
to favor motions with fewer geometric artifacts.
Replacing the physical tracking reward with a kinematic proxy gives a Success Rate of 77.98\%, below the pretrained model's 78.79\%.

Adding this reward to the joint fidelity and tracking rewards yields a comparable Success Rate (82.57\% versus 82.47\%), but higher reconstruction errors and a slightly lower Tracking Score.
We use the two-reward configuration for its better reconstruction at comparable tracking performance.

\begin{figure}[!tp]
  \centering
  \includegraphics[width=\linewidth]{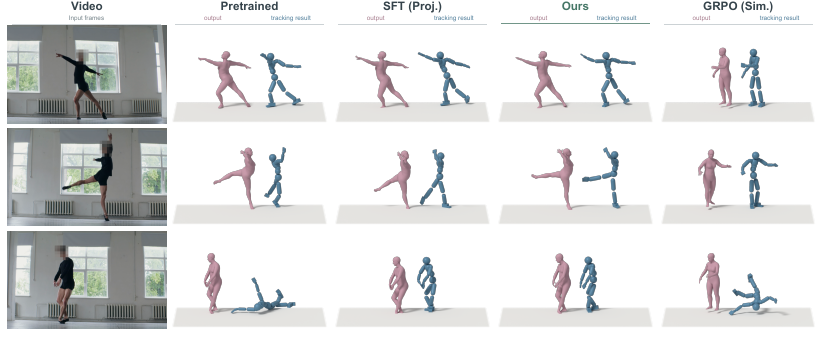}
  \caption{\textbf{Qualitative comparison of post-training strategies.} Columns show the input video, pretrained FlowHMR, SFT with projected targets (Proj.), joint-reward GRPO (Ours), and tracking-only GRPO (Sim.). Rows show three synchronized time points. Each method panel pairs the reconstructed output (pink, left) with its tracking result in simulation (blue, right).}
  \label{fig:posttraining_qualitative}
\end{figure}

\Needspace{9\baselineskip}
\paragraph{Qualitative comparison.}
\Cref{fig:posttraining_qualitative} illustrates the differences between post-training strategies.
The pretrained model fails during tracking and falls.
Projected-target SFT degrades reconstruction fidelity: in the second row, the reconstructed pose no longer matches the leg lift in the input frame.
Tracking-only GRPO exhibits reward hacking, keeping the character in place instead of reproducing the input motion.
Our joint-reward model better preserves the input action while maintaining upright tracking in this example.

\FloatBarrier

\subsection{Sampling Variability}
\label{app:sampling_analysis}

We evaluate fixed pretrained and final GRPO checkpoints with eight seeds on a 300-video subset of Wild-4K: 153 normal, 99 self-occluded, and 48 difficult motions. Table~\ref{tab:sampling_variability} reports variation among generated motions and execution quality under PHC+, using the metrics defined in Section~\ref{sec:setup}.

\begin{table}[!htbp]
\centering\small\setlength{\tabcolsep}{3pt}
\caption{\textbf{Sampling before and after post-training.} Ours w/o RL and Ours are each evaluated with eight seeds on a 300-video subset of Wild-4K. (a) Pairwise seed MPJPE averages the 28 pairs per video, then takes the median across videos. Tracking MPJPE averages the eight seed-level errors between execution and generated references over all motions. (b) Mean $\pm$ standard deviation of the eight seed-level PHC+ results.}
\label{tab:sampling_variability}
\begin{minipage}[t]{0.48\linewidth}
\centering
\textbf{(a) Motion variation and tracking}\par\smallskip
\begin{tabular}{@{}lrr@{}}
\toprule
Model & \makecell{Pairwise seed\\MPJPE (mm) $\downarrow$} & \makecell{Tracking MPJPE\\(cm) $\downarrow$} \\
\midrule
Ours w/o RL & 57.25 & \underline{8.35} \\
Ours & 22.86 & \textbf{6.96} \\
\bottomrule
\end{tabular}
\end{minipage}\hfill
\begin{minipage}[t]{0.48\linewidth}
\centering
\textbf{(b) Execution across seeds}\par\smallskip
\begin{tabular}{@{}lrr@{}}
\toprule
Model & \makecell{Success Rate (\%)\\$\uparrow$} & \makecell{Tracking Score\\$\uparrow$} \\
\midrule
Ours w/o RL & $\underline{76.33} \pm 1.02$ & $\underline{0.6799} \pm 0.0055$ \\
Ours & $\mathbf{79.71} \pm 0.42$ & $\mathbf{0.7344} \pm 0.0019$ \\
\bottomrule
\end{tabular}
\end{minipage}
\end{table}

\Needspace{6\baselineskip}
After GRPO, mean PHC+ Success Rate increases from 76.33\% to 79.71\%, while its standard deviation across eight seeds falls from 1.02 to 0.42 percentage points. Pairwise seed MPJPE decreases from 57.25 to 22.86\,mm, and tracking MPJPE decreases from 8.35 to 6.96\,cm. These results indicate better mean tracking performance and lower sensitivity to sampling seeds on this subset. Reduced variation across generated samples is distinct from the movement amplitude within a motion sequence.

\Needspace{34\baselineskip}
\subsection{Qualitative Comparisons}
\label{sec:qualitative}

\begin{figure}[!tp]
  \centering
  \includegraphics[width=\linewidth]{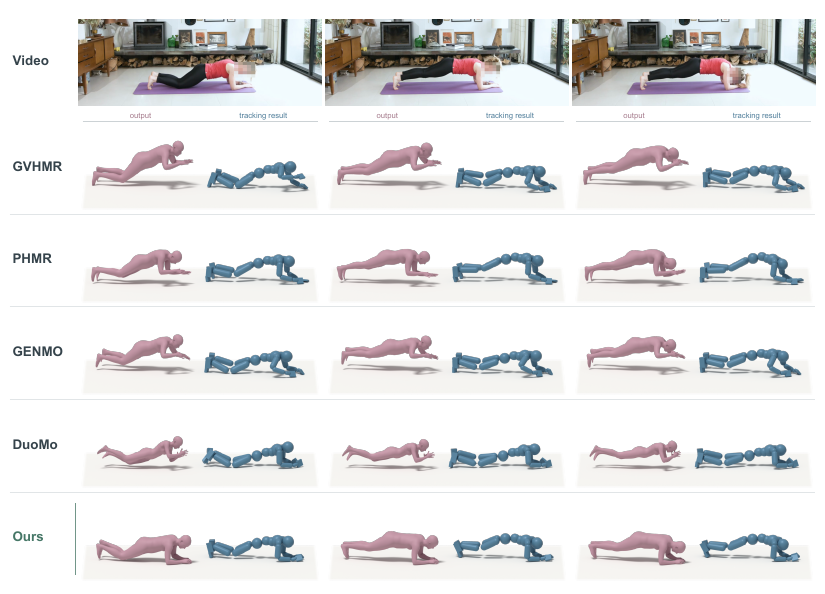}
  \caption{\textbf{Qualitative comparison.} The top row shows the input video, followed by GVHMR, PromptHMR (PHMR), GENMO, DuoMo, and Ours.}
  \label{fig:qualitative_extended}
\end{figure}

\begin{figure}[!tp]
  \centering
  \begin{minipage}{0.95\linewidth}
  \centering
  \includegraphics[width=\linewidth]{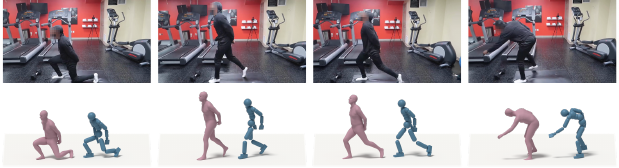}\par\vspace{8pt}
  \includegraphics[width=\linewidth]{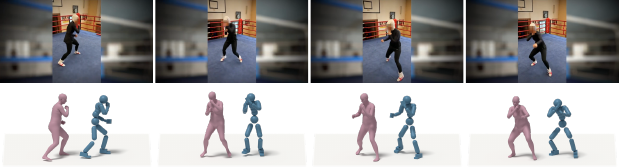}
  \end{minipage}
  \caption{\textbf{Additional qualitative examples.} We present two challenging cases: a fitness exercise that demands precise balance control, and a fast-paced boxing sequence. In both cases, our generated motions are successfully tracked in the physics simulator, demonstrating that our results remain physically balanced while faithfully following highly dynamic movements.}
  \label{fig:qualitative_examples}
\end{figure}

In \cref{fig:qualitative}, GVHMR~\citep{shen2024gvhmr} and DuoMo~\citep{wang2026duomo} fall during PHC+ tracking, while GENMO~\citep{li2025genmo} deviates substantially from its generated reference.
\modelname{} remains upright with torso orientation and arm poses closer to the reference. Visually plausible reconstruction alone does not guarantee stable physical execution.
Additional qualitative examples and comparisons appear in \cref{fig:qualitative_examples,fig:qualitative_extended,fig:posttraining_qualitative}.


\begingroup
\interlinepenalty=10000
\section{Limitations and Future Work}
\label{sec:limitations}

Physics-based simulation provides feedback that improves the physical plausibility and tracking of recovered motion.  However, physical execution is currently evaluated in a simplified environment containing only a ground plane.  Simulation-based post-training also requires repeated rollouts for sampled motions, making training costly as the data scale, sequence length, or number of samples grows.

Internet videos often contain interactions with objects and support surfaces.  Future work could reconstruct these entities and model their contacts in simulation, while developing more efficient simulation and rollout strategies to scale physical feedback to more videos and longer sequences.

\section{Conclusion}
\label{sec:conclusion}
We presented \modelname{}, a generative framework for recovering global 3D human motion from monocular video.  It combines Flow Matching pretraining with physics-based reinforcement learning to jointly optimize reconstruction fidelity and simulation tracking.  We also introduced Wild-4K and found improved overall motion quality and tracking performance over the evaluated baselines.

\par
\label{main:text-end}
\endgroup
\FloatBarrier

\clearpage
\bibliographystyle{flowhmr}
\bibliography{references,posttraining}
\end{document}